\documentclass[a4papper,review,sort&compress,preprint,12pt]{elsarticle}

\usepackage{amssymb}
\usepackage{amsmath}
\usepackage{booktabs}
\usepackage{multirow}
\usepackage{color, xcolor}
\usepackage{array}
\journal{Neurocomputing}

\usepackage{caption}
\begin{document}

\begin{frontmatter}

%% Title, authors and addresses

%% use the tnoteref command within \title for footnotes;
%% use the tnotetext command for theassociated footnote;
%% use the fnref command within \author or \address for footnotes;
%% use the fntext command for theassociated footnote;
%% use the corref command within \author for corresponding author footnotes;
%% use the cortext command for theassociated footnote;
%% use the ead command for the email address,
%% and the form \ead[url] for the home page:
%% \title{Title\tnoteref{label1}}
%% \tnotetext[label1]{}
%% \author{Name\corref{cor1}\fnref{label2}}
%% \ead{email address}
%% \ead[url]{home page}
%% \fntext[label2]{}
%% \cortext[cor1]{}
%% \affiliation{organization={},
%%             addressline={},
%%             city={},
%%             postcode={},
%%             state={},
%%             country={}}
%% \fntext[label3]{}

\title{A Multi-scale Information Integration Framework for Infrared and Visible Image Fusion}

% First author
%
% Options: Use if required
% eg: \author[1,3]{Author Name}[type=editor,
%       style=chinese,
%       auid=000,
%       bioid=1,
%       prefix=Sir,
%       orcid=0000-0000-0000-0000,
%       facebook=<facebook id>,
%       twitter=<twitter id>,
%       linkedin=<linkedin id>,
%       gplus=<gplus id>]
\author[1]{Guang Yang}
\ead{gyang2014@stu.xidian.edu.cn}

%  Credit authorship
% \credit{Conceptualization of this study, Methodology, Software}

% Second author
\author[1]{Jie Li}
\ead{leejie@mail.xidian.edu.cn}

% Third author
\author[1]{Hanxiao Lei}
\ead{21021210745@stu.xidian.edu.cn}

% Fourth author
\author[1,2]{Xinbo Gao \corref{cor1}}
\ead{xbgao@mail.xidian.edu.cn}

%\address[author1]{State Key Laboratory of Integrated Services Networks, School of Electronic Engineering, Xidian University, Xi'an, Shaanxi 710071, China}
%
%\address[author2]{Chongqing Key Laboratory of Image Cognition, Chongqing University of Posts and Telecommunications, Chongqing, Chongqing 400065, China}
\cortext[cor1]{Corresponding author}

% Address/affiliation
\affiliation[1]{organization={State Key Laboratory of Integrated Services Networks, School of Electronic Engineering, Xidian University},
    % addressline={Radarweg 29}, 
    city={Xi'an},
    % citysep={}, % Uncomment if no comma needed between city and postcode
    postcode={Shaanxi 710071}, 
    % state={},
    country={China}}

% Address/affiliation
\affiliation[2]{organization={Chongqing Key Laboratory of Image Cognition, Chongqing University of Posts and Telecommunications},
    % addressline={}, 
    city={Chongqing},
    % citysep={}, % Uncomment if no comma needed between city and postcode
    postcode={Chongqing 400065}, 
    % state={Trivandrum},
    country={China}}

\begin{abstract}
Infrared and visible image fusion aims at generating a fused image containing the intensity and detail information of source images, and the key issue is effectively measuring and integrating the complementary information of multi-modality images from the same scene. Existing methods mostly adopt a simple weight in the loss function to decide the information retention of each modality rather than adaptively measuring complementary information for different image pairs. In this study, we propose a multi-scale dual attention (MDA) framework for infrared and visible image fusion, which is designed to measure and integrate multi-scale complementary information in both structure and loss function at the image and patch level. In our method, the residual downsample block decomposes source images into three scales first. Then, dual attention fusion block integrates complementary information and generates a spatial and channel attention map at the same and adjacent scale for feature fusion. Finally, the output image is reconstructed by the residual reconstruction block. Loss function consists of image-level, feature-level and patch-level three parts, of which the calculation of the image-level and patch-level two parts are based on the weights generated by the complementary information measurement. Indeed, to constrain the pixel intensity distribution between the output and infrared image, a style loss is added. Our fusion results perform robust and informative across different scenarios. Qualitative and quantitative results on three datasets illustrate that our method is able to preserve both thermal radiation and detailed information from two modalities and achieve comparable results compared with the other state-of-the-art methods. Ablation experiments show the effectiveness of our information integration architecture and adaptively measure complementary information retention in the loss function. Our code is available at https://github.com/SSyangguang/MDA.
\end{abstract}


\end{frontmatter}

%% \linenumbers

%% main text
\section{Introduction}

Due to the limitations of different physical sensors, images from one single sensor is hard to comprehensively describe the scene information. Image fusion is one kind of multimodal information fusion technique, which aims at integrating images from different modalities into a single more informative image at the pixel level \cite{ma2019infrared, guo2019fusegan, xiang2022recognition}. Infrared and visible image fusion is a branch of the image fusion techniques to generate an image containing both thermal and detail information, in which the infrared images can easily distinguish targets with high temperature from the background \cite{zhang2020vifb}. However, infrared images typically have low spatial resolution and poor details. As complementary information to the infrared images, the visible images contain rich textures but are sensitive to unfavorable illumination conditions like nighttime and bad weather. The fusion results of infrared and visible images can depict the scene from aspects of thermal radiations and the human visual system simultaneously, which is practical in many fields, such as recognition, surveillance and object detection \cite{jian2020sedrfuse, muller2009cognitively, ma2020infrared}.

According to the adopted theories \cite{li2017pixel, kong2022guided, wang2014exposure}, the traditional image fusion method can be divided into several categories, namely multi-scale transform-, sparse representation-, subspace-, saliency-bases methods, hybrid models, and other models. However, the fusion performance of these methods is highly related to their handcrafted features and fusion schemes \cite{luo2021ifsepr}. In recent years, deep learning-based methods have been successfully applied to image fusion with their strong feature extraction ability and adaptive fusion strategy, which overcome the drawbacks of traditional methods to a certain extent. The overall deep learning-based image fusion pipeline can be processed implicitly by an end-to-end convolutional neural network under the well-designed network architecture and loss function, which is able to characterize complex relationships between input image pairs and the fusion image\cite{liu2018deep}. As suggested by \cite{zhang2021image}, the deep learning-based image fusion methods can be divided into autoencoder (AE)-based methods, convolutional neural network (CNN)-based methods and generative adversarial network (GAN)-based methods. Among them, GAN-based methods rely on an adversarial model between the generator and discriminator to implicitly fulfill feature extraction, feature fusion and image reconstruction \cite{ma2019fusiongan, song2022triple}. The AE-based and CNN-based methods generally use an autoencoder or fully convolutional network to realize the feature extraction and reconstruction, while the fusion strategy is conventional fusion rules or concatenation on channel dimension, such as DenseFuse \cite{li2018densefuse} and PMGI \cite{zhang2020rethinking}. 

Infrared and visible images capture radiation and illumination information respectively, and typically pixel intensity is used to constrain thermal radiation information retention while gradient variation is used to constrain texture detail information \cite{zhao2020learning}. Nevertheless, the complementary information varies greatly in different scenes. For instance, some infrared images achieve better detail at night, which makes the above information measurement method not accurate enough \cite{ma2020ganmcc}. Existing methods prefer preserving pixel intensity and detail information via the mean square error and structural similarity in the loss function separately, which did not consider their information retention-based adaptive weight coefficients before terms in the loss function. Most of them simply add residual or dense connection in the network architecture to capture texture details, failing to measure the multi-scale complementary information among different modalities and scales, which is a key issue in image fusion \cite{meher2019survey}.  Additionally, no ground truth in the infrared and visible image fusion task makes manually designed information retention methods unreasonable. Therefore, determining the degree of vital information retention based on each modality's own attributes and effectively integrating complementary information makes infrared and visible image fusion more reasonable. The existing methods barely applied complementary information measurement neither in calculating loss function nor feature fusion module, and the cooperation of the multi-scale complementary information between loss function and fusion structure is insufficient. U2Fusion \cite{xu2020u2fusion} adopted complementary information measurement in the loss function, but it did not consider measurement of the multi-scale information, such as the image-level and patch-level information capturing global and local complementary information.

\cite{lin2015remotely} showed that multi-scale analysis is similar to the human visual characteristics, and the fused image with such property is capable of capturing contextual information from the high-level representations and detailed information from the low-level representations \cite{zamir2020learning}. In this paper, we propose a multi-scale dual attention (MDA) framework applying multi-scale complementary information both in loss function and network architecture to effectively measure and integrate vital information at three scales. The given images are first decomposed into three scales via the residual downsample block. High-resolution features are more beneficial for measuring and preserving detailed information, while the pixel intensity distribution can be measured accurately at low resolution. Differing from those directly aggregate features by addition or concatenation, we introduce channel and spatial attention mechanisms to exploit 'what' and 'where' is informative for the given infrared-visible image pairs \cite{hu2018squeeze, chen2017sca}, and then fuse features extracted by the residual downsample block based on the attention maps. Dual attention fusion block merges features from the same and adjacent scale of different modalities along the spatial and channel dimensions, each of them generating an attention map for feature aggregation, which is considered the multi-scale complementary information interaction process. Meanwhile, each of the fusion blocks is constrained by the hierarchical loss to ensure the efficiency of the complementary information interaction and integration between the same and different scales. Fused features are upsampled to the same size as the input image and reconstructed by the residual reconstruction block.

As an auxiliary way to fuse the multi-scale complementary information in the network structure, the pixel item in the loss function consists of intensity and detail two parts with different adaptive weight coefficients obtained by the complementary information measurement. The generated coefficients are calculated based on the full resolution inputs and local patch, which is considered as the multi-scale complementary information assistance in the loss function. For the semantic-rich representations, we adopt statistical methods to generate weight coefficients for the pixel intensity term in the loss function. For the detail-rich representations,  we apply the gradient operator on the input image pairs to measure the retention of the texture information, and then weight coefficients of the detail term are generated.  The above weight generation method is applied to the image-level and patch-level simultaneously to assist multi-scale information in the network structure, enabling the multi-scale complementary information interaction efficiently. Additionally, we add a style loss to constrain pixel intensity distribution differences between the fusion image and the infrared image.
The main contributions in this work are as follows:
\begin{enumerate}[\textbullet]
\item We design a multi-scale dual attention (MDA) framework for infrared and visible image fusion that measures and fuses multi-scale complementary information from different modalities across multiple spatial scales, both utilizing pixel intensity and texture detail information. The network decomposes the image pairs into three scales to capture semantic-rich and detail-rich information, meanwhile the image-level and patch-level constraints are applied to assist the multi-scale complementary information integration.
\item A dual attention fusion block is designed for information integration based on spatial and channel attention mechanisms, where the vital spatial region and channel importance are determined by the attention map for multi-scale feature fusion. Feature maps of the same and adjacent scales are fused via the dual attention fusion block to ensure the efficiency of the multi-scale information interaction. 
\item Complementary information of image-level and patch-level infrared and visible representations is effectively measured through the statistical methods and gradient operator and generates adaptive weight coefficients for each item in the loss function to constrain the difference of fusion results and input image pairs, quantifying the degree of the vital information retention of thermal radiation and detail characteristics.
\item Extensive experiments are implemented on the \emph{TNO}, \emph{RoadScene} and $M^3FD$ datasets, illustrating that our complementary information measurement applied multi-scale network obtains competitive performances compared with seven state-of-the-art infrared and visible image fusion methods in terms of qualitative and quantitative comparisons. Ablation studies show the effectiveness of our multi-scale information measurement and integration method.
\end{enumerate}

The remainder of this article is organized as follows. Section II introduced previous related works. In Section III, we propose a multi-scale dual attention (MDA) framework for infrared and visible image fusion. Extensive experiments compared with seven methods on the \emph{TNO}, \emph{RoadScene} and $M^3FD$ datasets are illustrated in Section IV. Finally, Section V shows the conclusion.

\section{Related Work}

\subsection{Infrared and visible image fusion methods}
With the rapid development of deep learning, end-to-end image fusion methods applying neural networks have emerged, which can be divided into AE-based, CNN-based and GAN-based methods. Early AE-based methods \cite{li2018infrared, li2018densefuse} extracted features and reconstructed images through an encoder and decoder pretrained on the MS-COCO dataset, and integrated features via traditional fusion methods like addition, max selection and L1-norm fusion strategies. However, downsample operation in AE-based methods results in losing detailed information, meanwhile training on the MS-COCO dataset would make it difficult for the model to learn the property of infrared images. \cite{li2020nestfuse, li2021rfn} add short connections between the encoder and decoder to preserve detailed information. Part of the CNN-based methods \cite{li2021didfuse} adopted addition, weighted average and L1-norm to fuse features, while the others \cite{zhang2020rethinking,long2021rxdnfuse} concatenate features on the channel dimension to aggregate information. These kinds of methods fulfill feature extraction, fusion and reconstruction end-to-end without any downsample operations, and the fusion module learns to integrate modal-wise information through a well-designed loss function. FusionGAN \cite{ma2019fusiongan} firstly established an adversarial game between a generator and a discriminator to keep the thermal intensity information and visible detail information simultaneously, implementing feature extraction, fusion and reconstruction via the implicit way. However, the balance between the generator and a single discriminator is hard to balance during the training. DDcGAN \cite{ma2020ddcgan} and SDDGAN \cite{zhou2021semantic} adopted a generator to fuse the image and a dual-discriminator to force the fused image to retain intensity and detail information separately. AttentionFGAN \cite{li2020attentionfgan} and MFEIF\cite{liu2021learning} introduce the attention mechanism into the generator and discriminator to force the fusion results focusing on the most discriminator target regions. YDTR\cite{tang2022ydtr} extracted local features and significant context information via a Y-shape dynamic Transformer. SwinFusion \cite{ma2022swinfusion} devised a general image fusion framework based on the cross-domain long-range learning and Transformer. LRRNet \cite{li2023lrrnet} introduced learnable image decomposition to guide the construction of the network architecture.  FAFusion\cite{xiao2024fafusion} leveraged low- and high-frequency information for preserving the global brightness and local texture details of the infrared and visible images. Indeed, the multi-task image fusion framework has also been proposed, such as UMF-CMGR\cite{diunsupervised} combined cross-modality image alignment and fusion to avoid suffering ghosts of misaligned cross-modality image pairs. SeAFusion \cite{tang2022image} and PSFusion \cite{tang2023rethinking} combined image fusion tasks with high-level vision tasks to introduce semantic information from the semantic segmentation, which can improve the performance of both tasks simultaneously.

\subsection{Complementary information measurement}
VIF-Net \cite{hou2020vif} defined complementary information manually, i.e. considered the image intensity as the vital information and calculated the loss function based on the patch intensity. AUIF\cite{zhao2021efficient} decomposed images to base information and detail information as complementary information via a traditional optimization model. FusionDN \cite{xu2020fusiondn} and U2Fusion \cite{xu2020u2fusion} explored complementary information and applied it in loss function to achieve various image fusion tasks. EMFusion \cite{xu2021emfusion} defined complementary information as the unique channels of the pre-trained encoder and calculated the deep constraint part of the loss function based on these channels. CSF \cite{xu2021classification} quantified the contribution of each pixel to the classification result and defined it as complementary information, generating a pixel-level weight map to be the fusion strategy. DRF \cite{xu2021drf} and CUFD \cite{xu2022cufd} disentangled images into the common- and unique-related representations, and the unique-related representation is closer to each modality's individual sensor information, which could be considered the complementary information. STDFusionNet \cite{ma2021stdfusionnet} labels the saliency target mask to annotate regions humans prefer to pay attention to, and then combines the mask and a specific loss to guide the model retaining more thermal radiation information, which is considered as a manual complementation measurement method. TC-GAN\cite{yang2021infrared} generated a combined texture map to capture gradient changes.

\begin{figure*}[!t]
\centering
\includegraphics[width=5.5in]{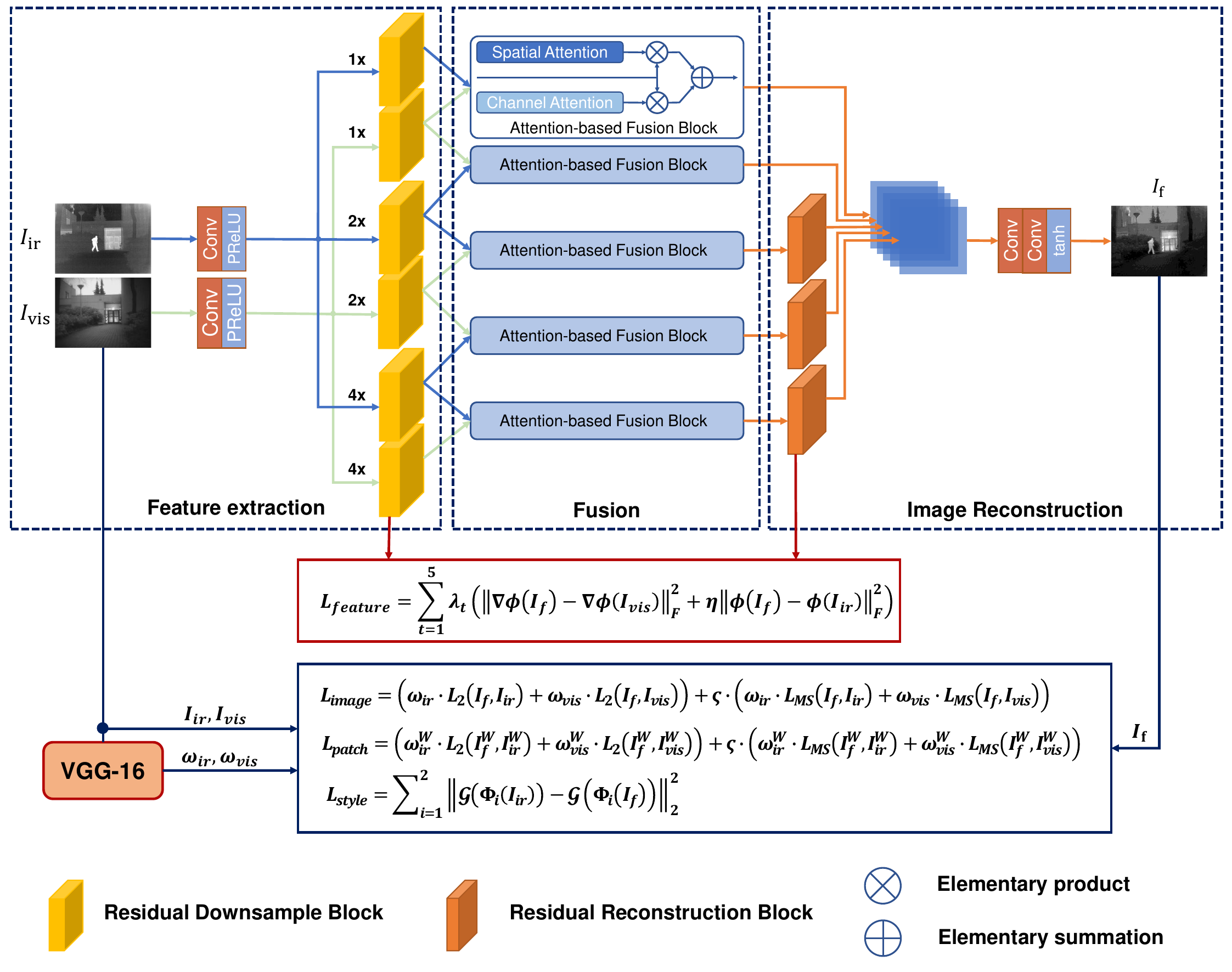}
\caption{
The pipeline of the proposed framework. The overall network is composed of two branches with infrared and visible images as inputs. Each branch is first decomposed into three scales by the residual downsample block. Then, features from two branches are integrated by dual attention fusion block and reconstructed by residual reconstruction block.
}
\label{fig:architecture}
\end{figure*}

\section{Methods}

In this section, we introduce the proposed MDA framework in detail. We introduce the overall framework first. Then, the architecture of dual attention fusion block combined spital and channel attention is described. Lastly, we define the complementary information measurement in the loss function.

\subsection{Network Architecture}

In this section, we present the network architecture of the proposed framework, as shown in Fig. \ref{fig:architecture}. As can be seen, our method contains three main parts: 1. Residual downsample block applies multi-resolution residual convolution streams for extracting both detail-rich and semantic-rich feature representations, 2. The spatial and channel attention-based dual attention fusion block aggregates features from both the same and adjacent scales, 3. Residual reconstruction block for image reconstruction by using the multi-scale fused features. The network as a whole is a two-branch structure, extracting multi-scale features of infrared and visible images respectively.

\begin{figure}[!t]
\centering
\includegraphics[width=5.5in]{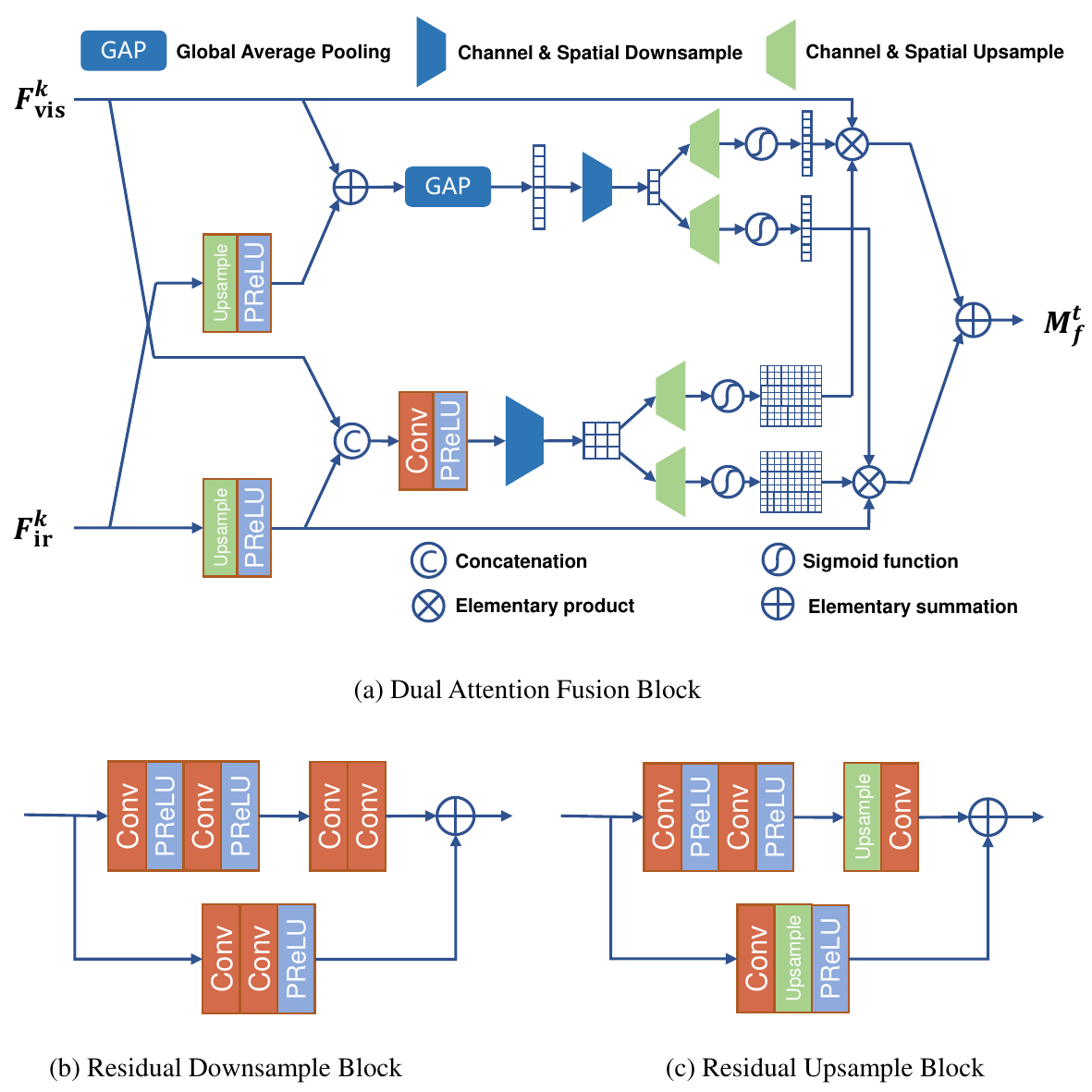}
\caption{
Illustration of the dual attention fusion block, residual downsample block and residual reconstruction block. Dual attention fusion block generates attention maps for each branch and outputs a fused feature map. Residual downsample/reconstruction block is performed to extract features and reconstruct images.
}
\label{fig:fusion_conv}
\end{figure}

Given the source images $ I_{ir} \in \mathbb{R}^{H \times W \times 1} $ and $ I_{vis} \in \mathbb{R}^{H \times W \times 3} $, we first convert the color space of visible image $ I_{vis} $ from RGB to YCbCr, and separate the Y channel to fuse with infrared image $ I_{ir} $ as well as calculating loss function. We use a single 3×3 convolutional layer with PReLU activation to extract basic features $ F_{ir},F_{vis} \in \mathbb{R}^{H \times W \times C} $, and expand the channel dimension from 1 to 32. Residual downsample block consists of three fully convolutional downsample streams, as is shown in Fig. \ref{fig:fusion_conv}b, extracting deep features $ F_{ir}^1,F_{vis}^1 \in \mathbb{R}^{H \times W \times C} $ ,$ F_{ir}^{2},F_{vis}^{2} \in \mathbb{R}^{\frac{H}{2} \times \frac{W}{2} \times C} $ and $ F_{ir}^{3},F_{vis}^{3} \in R^{\frac{H}{4} \times \frac{W}{4} \times C} $ via convolutional layers with kernel stride=1, 2, 4 respectively. After the feature extraction, the multi-scale feature representation preserves rich detail information by maintaining high-resolution feature maps, while conveying contextual information from the low-resolution feature maps.

Dual attention fusion block contains five individual fusion blocks in total, and three of them integrate features on their respective scales, while the other two integrate features from adjacent scales since the large resolution difference easily leads to noise, fusion difficulties and computational cost \cite{pang2020multi}. More specifically, the fusion between adjacent scales based on the high-resolution visible branch $ F_{vis}^k $ and low-resolution infrared branch $ F_{ir}^{k+1} $, for the reason that visible images generally have richer detail information. Fusion between the same and adjacent scales enables the information integration process to take the modality-specific and scale-specific complementary information into account simultaneously.

Fused features produced by five fusion blocks are at three scales, where the first two $ M^1_f,M^2_f \in \mathbb{R}^{H \times W \times C} $ keep the same resolution with the input. The middle two  $ M^3_f,M^4_f \in \mathbb{R}^{\frac{H}{2} \times \frac{W}{2} \times C} $ and the last one $ M^5_f \in \mathbb{R}^{\frac{H}{4} \times \frac{W}{4} \times C} $ are the half and quarter resolution of the input, respectively. Before concatenating multi-scale features from five fusion blocks, we use a residual upsample convolutional module to unify the resolution same with the input image pair $ H \times W $, as is shown in Fig. \ref{fig:fusion_conv}c. The output of the network $ I_{fusion} \in \mathbb{R}^{H \times W \times 1} $ is a grayscale fused image reconstructed by two 3×3 convolutional layers with tanh activation.

Finally, the fused illumination component was stacked with the Cb and Cr channels to produce the final fused image.

\subsection{Dual Attention Fusion Block}
Inspired by \cite{li2019selective,hu2018squeeze}, dual attention fusion block is composed of spatial attention and channel attention branches, and shares the same network architecture. With the application of attention, the redundant information in spatial and channel dimensions could be suppressed while the vital information could be enforced \cite{woo2018cbam}. Spatial and channel attention branches generate attention maps in parallel, i.e. the infrared and visible features are fed into the attention module simultaneously and generate attention maps for the two modalities. Therefore, we exploit the modal-wise feature representations and integrate multi-modality complementary information jointly rather than yielding attention maps for each modality on the channel dimension independently.

Taking the highest spatial resolution feature maps $ F_{ir}^k $,$ F_{vis}^k \in \mathbb{R}^{H \times W \times C} $ as an example, channel attention module exploits the inter-channel relationship between different modalities. As shown in Fig. \ref{fig:fusion_conv}a, the low-resolution feature map is first upsampled to the same size as the larger resolution feature map, and then complementary information from two branches is aggregated via the element-wise summation and global average pooling (GAP), squeezing the 2D spatial features to a 1D channel descriptor $ {F_{CM}} \in \mathbb{R}^{1 \times 1 \times C} $. To further improve efficiency, the dimension of $ F_{CM} $ is reduced by a fully connected layer and then expanded to $ C $ by a dual fully connected layer, which corresponds to the two input branches, respectively. The above generation of channel attention map is as follows:
\begin{equation}
  {CM}_{ir}^k, {CM}_{vis}^k = \sigma (MLP(GAP(F_{ir}^k+F_{vis}^k)))
\end{equation}

\noindent where $MLP$ denotes two fully connected layers, and $\sigma$ denotes sigmoid function. The above operation enables the fusion features to adaptively select the inter-channel complementary information.

%\begin{figure}[!t]
%\centering
%\subfloat[Dual Attention Fusion Block]{\includegraphics[width=85mm]{fusion}%
%\label{fig:Dual Attention Fusion Block}}
%\hfil
%\subfloat[Residual Downsample Block]{\includegraphics[width=38mm]{residual_down}%
%\label{fig:Residual Downsample Block}}
%\hfil
%\subfloat[\tiny{Residual Reconstruction Block}]{\includegraphics[width=38mm]{residual_up}%
%\label{fig:Residual Reconstruction Block}}
%
%\caption{Illustration of the dual attention fusion block, residual downsample block and residual reconstruction block. Dual attention fusion block generates attention maps for each branch and outputs a fused feature map. Residual downsample/reconstruction block is performed to extract features and reconstruct images.}
%\label{fig:fusion_conv}
%\end{figure}

Spatial attention module is designed to exploit the inter-spatial relationship between two modalities, and similar to the channel attention module in structure, the difference is that the spatial attention module first concatenates feature maps from two branches rather than element-wise summation and global average pooling. Then, convolutional streams successively reduce the spatial dimension and yield spatial attention maps $ {SM}_{ir}^k, {SM}_{vis}^k \in \mathbb{R}^{H \times W \times C} $ for two input branches, respectively. The generation of the spatial attention map is as follows:

\begin{equation}
  {SM}_{ir}^k, {SM}_{vis}^k = \sigma (f^{3 \times 3}(concat(F_{ir}^k, F_{vis}^k)))
\end{equation}

\noindent where $ f^{3 \times 3} $ denotes a downsample convolutional stream with $3 \times 3$ kernel size.

After the attention block infers a 1D channel attention map $ {CM}_{ir}^k , {CM}_{vis}^k \in \mathbb{R}^{1 \times 1 \times C} $ and a 2D spatial attention map $ {SM}_{ir}^k , {SM}_{vis}^k \in \mathbb{R}^{H \times W \times C} $, the fusion feature map is composed as:

\begin{equation}
  M^k_f={CM}_{ir}^k \otimes ({SM}_{ir}^k \otimes F_{ir}^k )+{CM}_{vis}^k \otimes ({SM}_{vis}^k \otimes F_{vis}^k )
\end{equation}

\noindent where $ k $ denotes the $ k $-th scale of input feature maps, and $ F_{ir}^k, F_{vis}^k $ and $  M^k_f $ are input feature maps of infrared, visible and the output feature maps, respectively. $\otimes$ denotes element-wise multiplication and the channel attention maps are broadcasted along the spatial dimension during the multiplication.

\subsection{Loss Function}
Items related to the visible images in the loss function are only based on the illumination component, namely the Y channel of the visible images. Since our work aims to effectively measure and preserve complementary information between modalities, the loss function is designed to measure the vital information of each modality first, and then constrain the output based on the vital information calculation results. Specifically, we apply an information measurement method to estimate vital information and generate a weight for the pixel item in the loss function, determining the information retention degree. Hierarchical loss is added to guarantee the training efficiency of the dual attention fusion block \cite{deng2021deep}. Besides, \cite{li2017demystifying} argues that the essence of neural style transfer is to match the feature distributions between the style images and the generated images. Since the pixel grayscale distribution of infrared images is highly correlated with thermal radiation, we expect fusion images and infrared images to be closer in style.
The total loss function of our method is defined as follows:
\begin{equation}
  L_{total} = L_{pixel} + \alpha \cdot L_{feature} + \beta \cdot L_{style}
\end{equation}

\noindent where $ \alpha $ and $ \beta $ controls the balance of $ L_{pixel} $, $ L_{feature} $ and $ L_{style} $. These three items are described in detail below.

\begin{figure}
\centering
\includegraphics[width=5in]{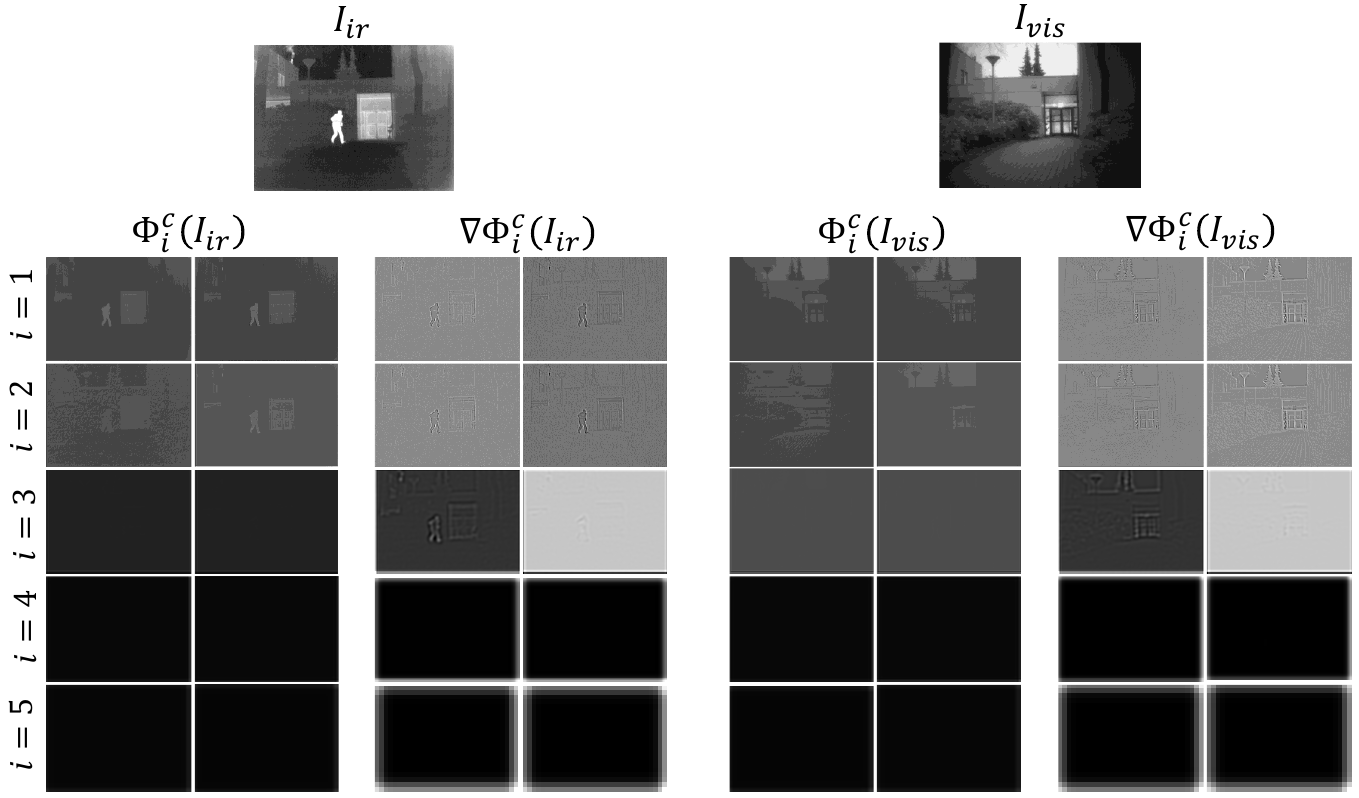}
\caption{
Visualization of feature maps extracted by VGG-16. Given an infrared and visible image pair, from up to bottom are feature maps extracted by convolutional layers before five max-pooling layers. $ i $ denotes feature maps extracted before the $ i $-th max-pooling layer of VGG-16.
}
\label{fig:vgg}
\end{figure}

\cite{johnson2016perceptual} indicates that based on high-level features extracted from a pre-trained network, perceptual loss measures image similarities more robustly than pixel loss during the training, thus we adopt the pre-trained VGG-16 network \cite{simonyan2014very} to measure the complementary information and generate weights for each item in the loss function. However, we find that deep features of VGG-16 barely convey intensity information when our small-size grayscale images as the input. As we can see in Fig. \ref{fig:vgg}, feature maps in the deep layers of VGG-16 contain poor information, such as the fourth and fifth layers. Considering the precision of complementary information measurement, we estimate the weights in the loss function only based on shallow convolutional layers before the first two max-pooling layers of the VGG-16 network.

Entropy is considered the way to measure the amount of information contained in an image \cite{roberts2008assessment} and standard deviation reflects the contrast, which can be used to measure the overall pixel intensity distribution of an image. Based on the above considerations, the weights to constrain intensity item is designed as follows:

\begin{equation}
  \omega_{int} = \frac{1}{2} \sum_{i=1}^{2} \bigg( \frac{1}{C} \sum_{c=1}^{C} \Big( \text{en} \big( \Phi_{i}^{c} \left( I \right) \big) + \delta \cdot \text{std} \big( \Phi_{i}^{c} \left( I \right) \big) \Big) \bigg)
\end{equation}

\begin{equation}
\label{eq:int_soft}
  \omega_{int}^{ir}, \omega_{int}^{vis} = \text{softmax} \left(  \frac{\omega_{int}^{ir}}{c_{int}} , \frac{\omega_{int}^{vis}}{c_{int}}  \right)
\end{equation}

\noindent where $ \text{en}() $ and $ \text{std}() $ are entropy and standard deviation operator for the whole image, and $ \delta $ is a weight to control the trade-off between the entropy and standard deviation. $ \Phi_{i}^{c} \left( I \right) $ is a feature map by the convolutional layers before the $ i $-th max-pooling layer of the $ I $, and $ c $ denotes the $ c $-th channels. We use the softmax function to keep the sum of the $ \omega_{int}^{ir} $ and $ \omega_{int}^{vis} $ is 1.

Meanwhile, the detail information of images, especially visible images, can be characterized by gradient variants, hence we utilize the L2 norm of the image gradient to measure the detail texture information. Due to the sensitivity to noise when used along, Laplacian smoothed with Gaussian smoothing filter, i.e. Laplacian of Gaussian (LoG) is applied to use for edge detection \cite{fang2021superpixel}. Therefore, the gradient item is designed as follows:

\begin{equation}
    \omega_{grad} = \frac{1}{2} \sum_{i=1}^{2} \Big( \frac{1}{C} \sum_{c=1}^{C} \big(  \left \| \nabla  \Phi_{i}^{c} \left( I \right) \right \|_2^2  \big) \Big)
\end{equation}

\begin{equation}
\label{eq:grad_soft}
  \omega_{grad}^{ir}, \omega_{grad}^{vis} = \text{softmax} \left(  \frac{\omega_{grad}^{ir}}{c_{grad}} , \frac{\omega_{grad}^{vis}}{c_{grad}}  \right)
\end{equation}

\noindent where $ \nabla $ denotes the LoG gradient operator. $ \left \| \cdot \right \|_2 $ denotes the L2 norm. We also use the softmax function to keep the sum of the $ \omega_{grad}^{ir} $ and $ \omega_{grad}^{vis} $ equal to 1. Finally, $ \omega_{int} $ and $ \omega_{grad} $ are utilized to control the information retention degree of intensity items and gradient items, respectively.

After obtaining the weights for intensity and gradient items, we constrain the pixel difference between the input image pairs and the fusion result. For the gradient item, we adopt the mean structural similarity index metric (MSSIM) \cite{wang2004image} loss to constrain the texture difference between fusion result and input images, which can describe the perceptual quality of the image better than L2 loss. Then, $ L_{image} $ is defined as follows:

\begin{equation}
\begin{aligned}
    L_{image} = & \Big(  \omega_{grad}^{ir} \cdot \big(  1-MSSIM \left( I_f,I_{ir} \right)  \big) 
\\
& + \omega_{grad}^{vis} \cdot \big(  1-MSSIM \left( I_f,I_{vis} \right)  \big)  \Big)  
\\
& + \zeta \cdot \left(  \omega_{int}^{ir} \cdot \left \|  I_f-I_{ir}  \right \|_2^2  +  \omega_{int}^{vis} \cdot \left \|  I_f-I_{vis}  \right \|_2^2 \right)
\end{aligned}
\end{equation}

\noindent where $ MSSIM(I_x,I_y ) $ denotes the mean structural similarity between $ I_x $ and $ I_y $, and $ \zeta $ stands for the parameter controlling the trade-off between the gradient item and the intensity item.

Since the richer thermal radiation information of infrared images focuses on the area with a larger gray value, we expect that the pixel intensity of fusion result on that local area is similar to the corresponding position on infrared images, and so is the local texture detail of visible images. Hence we add another patch-wise loss in the pixel item to measure local complementary information, which can be described as follows.

\begin{equation}
\begin{aligned}
    L_{patch} & =  \Big( \omega_{grad}^{ir_W} \cdot \big(  1-SSIM \left( I_f^W,I_{ir}^W \right)  \big) 
\\
& + \omega_{grad}^{vis_W} \cdot \big(  1-SSIM \left( I_f^W,I_{vis}^W \right)  \big)  \Big)  
\\
& + \zeta \cdot \left(  \omega_{int}^{ir_W} \cdot \left \|  I_f^W - I_{ir}^W  \right \|_2^2  +  \omega_{int}^{vis_W} \cdot \left \|  I_f^W - I_{vis}^W  \right \|_2^2 \right)
\end{aligned}
\end{equation}

$ I_f^W $, $ I_{ir}^W $ and $ I_{vis}^W $ denote the fusion image, infrared image and visible image in the local sliding window $ W $, respectively. Considering the area of targets with high temperatures existing at boundaries between patches, we set the size of the sliding window to 21 ×21 and the overlap pixel between the sliding window is set to 1 in this study. Pixel intensity in a small area is mainly dominated by the gray value rather than entropy, thus we replace entropy with the mean pixel value in the window, and the weight calculation of patch loss is as follows:

\begin{equation}
\begin{aligned}
\omega_{int}^W & = \frac{\text{mean} (I^W) + \delta \cdot \text{std} (I^W)} {c_{int}} \\
\omega_{grad}^W & = \frac{ \left \| \nabla  I^W  \right \|_F^2} {c_{grad}}
\end{aligned}
\end{equation}

As in (\ref{eq:int_soft})(\ref{eq:grad_soft}), $ \omega_{int_W}^{ir} $, $ \omega_{int_W}^{vis} $, $ \omega_{grad_W}^{vis} $ and $ \omega_{grad_W}^{vis} $ are also eventually obtained by the softmax function keeping the sum of weight coefficient is 1. The pixel loss is the weighted sum of $ L_{image} $ and $ L_{patch} $, as shown in (\ref{eq:pixel}).

\begin{equation}
\label{eq:pixel}
  L_{pixel} = L_{image} + \gamma L_{patch}
\end{equation}

\noindent where $ \gamma $ controls the trade-off. Except for the constraint between input and fusion result, we also set hierarchical loss for each fusion block to guarantee the training efficiency of the dual attention module and multi-scale information interaction. Therefore, $ L_{feature} $ is defined as follows:

\begin{equation}
  L_{feature} = \sum_{t=1}^5 \lambda_t \big(  \left \|  \nabla  M_f^t - \nabla F_{vis}^t  \right \|_F^2 + \eta \cdot \left \|    M_f^t -  F_{ir}^t  \right \|_F^2  \big)
\end{equation}

\noindent where $ \lambda_t $ are weights to the constraint of each fusion block and $ \eta $ controlling the trade-off. $ \nabla $ denotes the LoG gradient operator.

Finally, the standard style loss is added to constrain the gray value distribution of infrared and fusion images, as shown in (\ref{eq:style}).

\begin{equation}
\label{eq:style}
  L_{style} = \sum_{i=1}^{2} {\left \| {\mathcal{G} \big( \Phi_{i} \left( I_{ir} \right) \big)} - {\mathcal{G} \big( \Phi_{i} \left( I_f \right) \big) } \right \|}_2^2
\end{equation}

\noindent $ \mathcal{G} $ is Gram matrix in \cite{gatys2016image} to represent feature correlations, which is the inner product between the vectorized feature maps. As the weight coefficient calculation in the pixel loss, we calculate style loss only based on feature maps before the first two max-pooling layers of the VGG-16 network, i.e. $ i=1,2 $.

\subsection{Implementations}

\subsubsection{Data}In the training phase, we use KAIST \cite{hwang2015multispectral} dataset to train our network, which consists of nearly 95k aligned IR-VIS image pairs (640×480) taken from the vehicle. We crop source images to patches of size 192 × 192 and transfer the color space of visible images to YCbCr. Since the infrared images are grayscale, we only use the Y channel to fuse with infrared images. In the testing phase, \emph{TNO} \footnote{https://figshare.com/articles/TNO\_Image\_Fusion\_Dataset/1008029.} dataset \cite{toet2014tno}, \emph{RoadScene}\footnote{https://github.com/hanna-xu/RoadScene.} dataset \cite{xu2020fusiondn} and $M^3FD$\footnote{https://github.com/JinyuanLiu-CV/TarDAL.} dataset provided by \cite{liu2022target} is used to evaluate the performance of our method. \emph{TNO} is a classical IR-VIS dataset with rich scenes. \emph{RoadScene} contains 221 aligned IR and VIS image pairs, which are highly representative scenes from the FLIR video, and those image pairs are aligned accurately while denoised. $M^3FD$ is a high-resolution IR-VIS dataset consisting of 4200 aligned pairs with six classes of annotated objects, and 300 pairs are chosen to evaluate the performance of methods.

\subsubsection{Training details} The batch size and epoch are set to 24 and 5, respectively. We set $ \alpha=1e-8 $, $ \beta=1e5 $, $ \gamma=2 $ and $ \eta=0.02 $ in the loss functions, respectively. For the complementary information measurement parameters in $ L_{image} $, the $ \delta $, $ \zeta $ and $ c $ is set as $ 0.167 $, $ 20 $ and $ 3 \times 10^3 $, respectively. We optimize the loss function using the Adam with a learning rate $ 1 \times 10^{-5} $. Experiments are performed on NVIDIA GeForce RTX 3090 GPU and 2.90GHz Intel Xeon Gold 6226R CPU.

\section{Experiments}

\subsection{Implementation}
In this section, we verify the effectiveness of our method through qualitative and quantitative evaluations by comparing it with several state-of-the-art methods on the \emph{TNO}, \emph{RoadScene} and $M^3FD$ datasets. These comparative methods include the CNN-based method, GAN-based methods, AE-based and Transformer-based methods, which are DenseFuse \cite{li2018densefuse}, RFN-Nest \cite{li2021rfn}, GANMcC \cite{ma2020ganmcc},  UMF-CMGR\cite{diunsupervised}, SwinFusion\cite{ma2022swinfusion}, LRRNet\cite{li2023lrrnet} and FAFusion\cite{xiao2024fafusion}. Qualitiative evaluation is based on the human visual system, hoping to highlight the thermal target throughout the scene while preserving texture detail. Quantitative evaluation access fusion image quality the way of several statistics and visual metrics.  We employ eight metrics to evaluate the quality of the fusion results, which are entropy (EN) \cite{roberts2008assessment}, visual information fidelity (VIF) \cite{han2013new}, the sum of the correlations of differences (SCD) \cite{aslantas2015new}, mean squared error (MSE), average gradient (AG)\cite{cui2015detail}, correlation coefficient (CC),  edge retentiveness ($ Q^{AB/F} $) \cite{petrovic2004evaluation} and standard deviation (SD) \cite{aslantas2015new}. EN measures the amount of information contained in an image based on the information theory. SD reflects the contrast of the fusion image. VIF measures the information fidelity of the fused image. MSE is the mean error value between the fusion image between source images. AG quantifies the gradient information. $ Q^{AB/F} $ measures the amount of edge information transferred from source images to the fused image. SCD and CC compute the degree of correlation degree between source images and the fusion result. Furthermore, as described in Section 3.3, for our small-size image pairs, deep features of the VGG-16 network contain poor intensity and detail information, so we investigate the performance of different depth VGG-16 features used in the loss function. Lastly, we visualize the spatial attention maps in the dual attention fusion block to validate the ability of information integration.

\begin{figure*}[!t]
\centering
\includegraphics[width=5.5in]{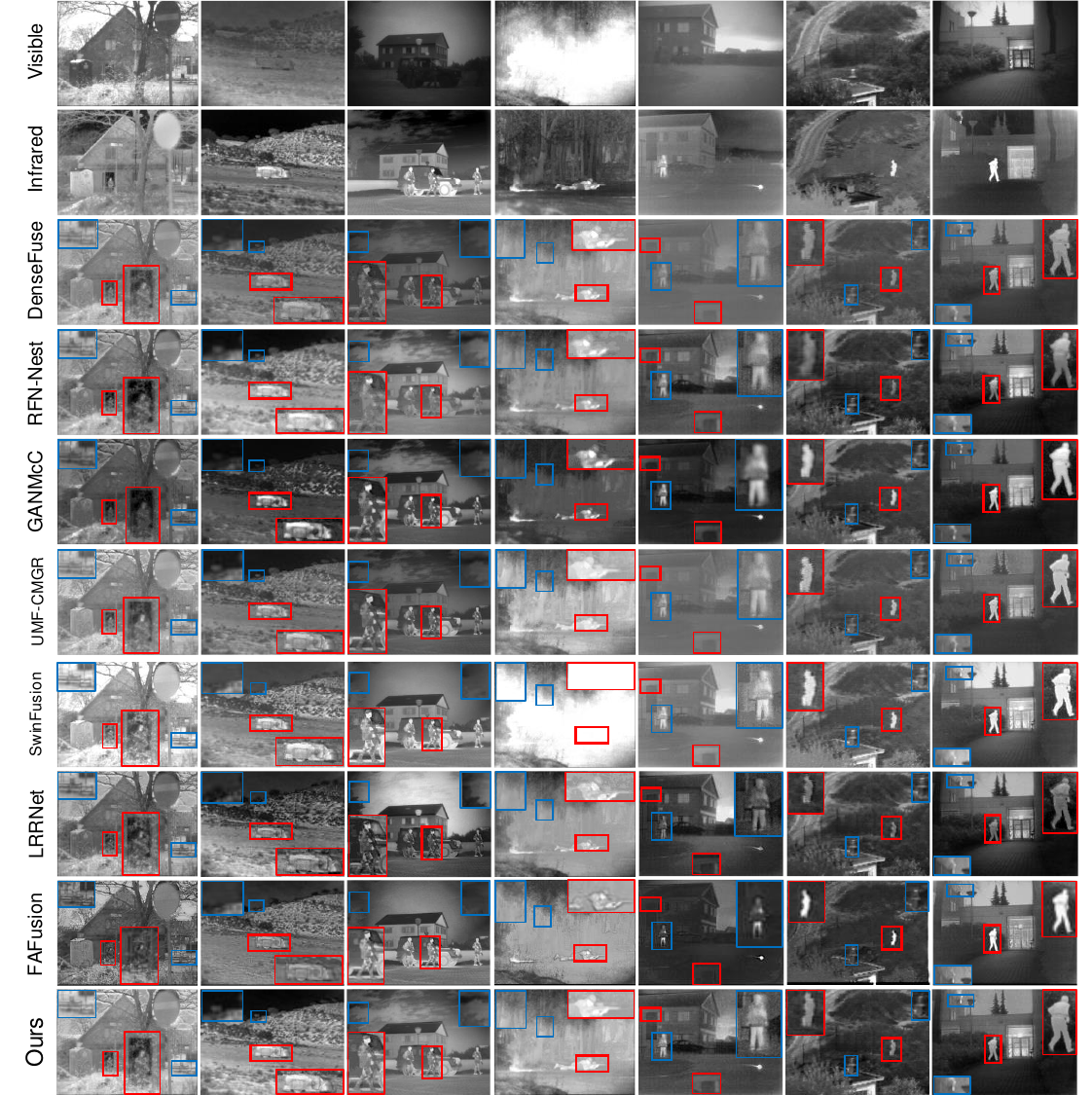}
\caption{
Qualitative comparison of our method with 7 state-of-the-art models on seven infrared and visible image pairs in the \emph{TNO} dataset. The first and second rows are visible and infrared images, respectively. From the third to the last rows are fusion results of DenseFuse\cite{li2018densefuse}, RFN-Nest\cite{li2021rfn}, GANMcC\cite{ma2020ganmcc}, UMF-CMGR\cite{diunsupervised}, SwinFusion\cite{ma2022swinfusion}, LRRNet\cite{li2023lrrnet}, FAFusion\cite{xiao2024fafusion} and our MDA.
}
\label{fig:model_comparison}
\end{figure*}

%------------------------------------------------------------------------- 
\begin{figure*}[!t]
\centering
\includegraphics[width=5.5in]{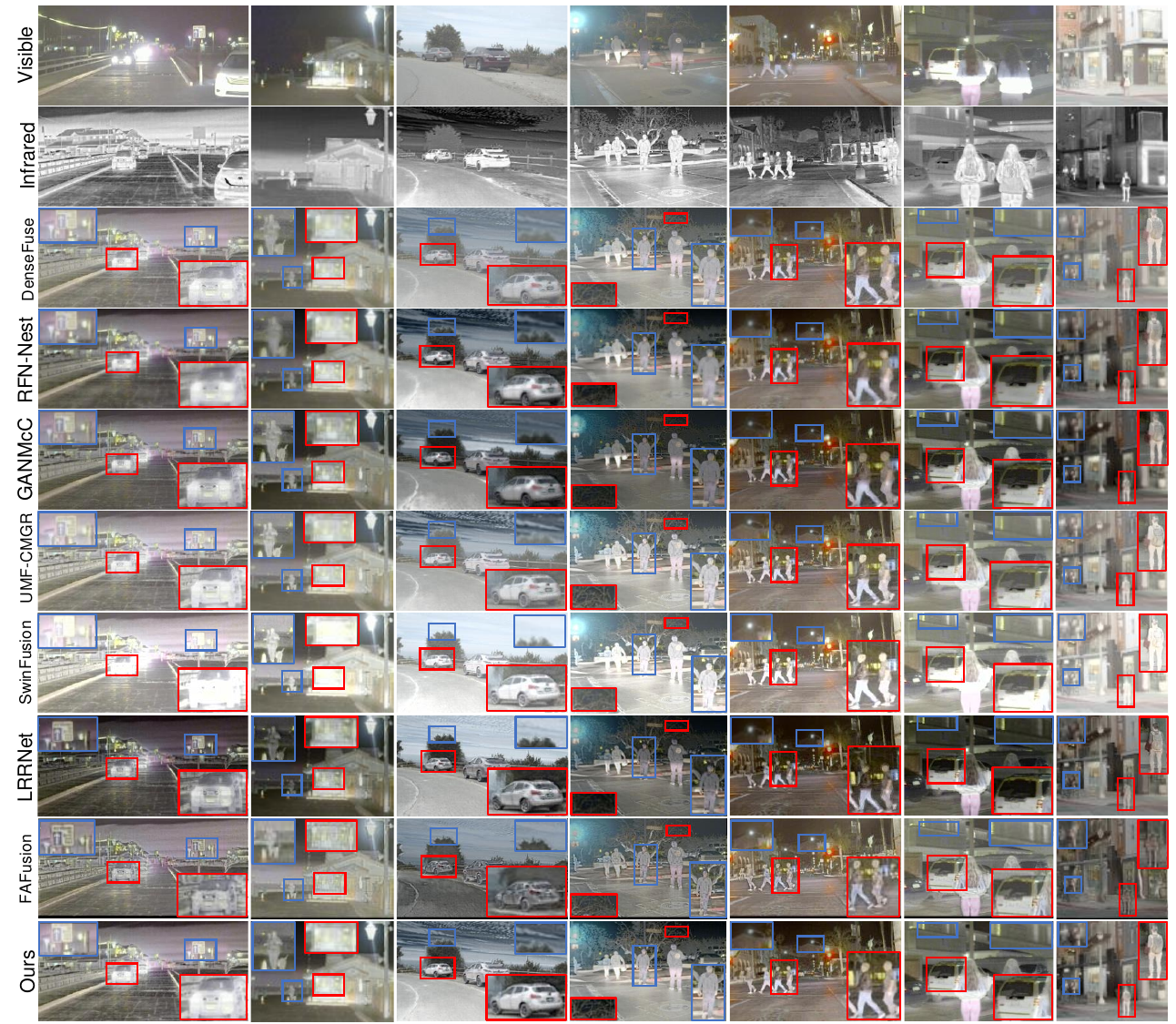}
\caption{
Qualitative comparison of our method with 7 state-of-the-art models on seven infrared and visible image pairs in the \emph{RoadScene} dataset. The first and second rows are visible and infrared images, respectively. From the third to the last rows are fusion results of DenseFuse\cite{li2018densefuse}, RFN-Nest\cite{li2021rfn}, GANMcC\cite{ma2020ganmcc}, UMF-CMGR\cite{diunsupervised}, SwinFusion\cite{ma2022swinfusion}, LRRNet\cite{li2023lrrnet}, FAFusion\cite{xiao2024fafusion} and our MDA.
}
\label{fig:model_comparison_roadscene}
\end{figure*}

\subsection{Results on TNO Dataset}

We conduct our method and the state-of-the-art methods on the \emph{TNO} dataset to validate fusion performance in terms of infrared images and visible images with noise.

\subsubsection{Qualitative Evaluation}

Fig. \ref{fig:model_comparison} exhibits qualitative comparison results of different models on the \emph{TNO} dataset. The top two rows and the last row represent the visible images, infrared images and fusion results of our method, respectively. From the third to the ninth row are fusion results of DenseFuse, RFN-Nest, GANMcC, UMF-CMGR, SwinFusion, LRRNet and FAFusion. First of all, as the purpose of MDA is to effectively measure and integrate the complementary information, we mainly follow whether both the thermal radiation information of the infrared image and the detailed information of the visible image are effectively retained. Secondly, style loss can keep the fusion result maintaining the same pixel intensity distribution as the infrared image. Lastly, the application of the multi-scale framework and patch-wise loss function drives the model to measure complementary information not only for the whole image but also for each local region.

Visual inspection shows that our results can preserve the high contrast of pixel intensity like the brighter targets in infrared images or illuminated regions in visible images. Meanwhile, our results preserve rich texture details from visible images and regions with dramatic pixel intensity changes in infrared images, such as the first column of Fig. \ref{fig:model_comparison}. As is shown in the red box of this image pair, a person standing near the door with a higher temperature is highlighted well as the infrared image, and the chair which is only can be seen in the visible image is also retained. Our goal is to adaptively measure and integrate complementary information from both modalities, therefore we can see from the fourth column that the branch with complex texture of the infrared image is preserved well, and so does the crawling person with higher pixel intensity. In addition, our results have fewer artifacts at the edge of regions where infrared and visible pixel intensity distributions collide compared with other methods, as can be seen from the reb boxes of the fourth, fifth and sixth columns in Fig. \ref{fig:model_comparison}, where the edge region of people and background keeps sharp.

The fusion result of the sixth and last column in Fig. \ref{fig:model_comparison} illustrates the superiority of our multiscale framework and patch-wise loss function, where both thermal and detail information in the region of running person is well preserved, such as the junction of clothes and pants. Other methods like GANMcC and SwinFusion prefer to retain the global pixel intensity of the infrared image, while the fusion result of RFN-Nest and LRRNet in the local region lacks thermal saliency. Thus, our method is able to measure and integrate vital information both in global and local areas.

\begin{table}[htbp]
  \centering
  \caption{Quantitative comparison of our method with 7 state-of-the-art models on the \emph{TNO}, \emph{RoadScene} and $M^3FD$ dataset. The best performance are shown in \textbf{bold}, and the second and third best performance are shown in \textcolor[rgb]{ 1,  0,  0}{red} and \textcolor[rgb]{ 0,  0,  1}{blue}, respectively.}
  \renewcommand{\arraystretch}{0.7}
  \resizebox{\textwidth}{!}{
  \setlength{\tabcolsep}{3mm}{
    \begin{tabular}{l|cccccccc}
    \toprule
    \multicolumn{1}{c}{\multirow{2}[4]{*}{Method}} & \multicolumn{8}{c}{$TNO$ Dataset} \\
\cmidrule{2-9}    \multicolumn{1}{c}{} & EN$\uparrow$  & VIF$\uparrow$ & SCD$\uparrow$ & MSE$\downarrow$ & AG$\uparrow$  & CC$\uparrow$  & $Q^{AB/F}$$\uparrow$& SD$\uparrow$ \\
    \midrule
    DenseFuse & 6.984  & 0.826  & \textcolor[rgb]{ 1,  0,  0}{1.829}  & \textcolor[rgb]{ 1,  0,  0}{0.043}  & 3.475  & 0.545  & 0.271  & \textcolor[rgb]{ 0,  0,  1}{9.679}  \\
    RFN-Nest   & \textcolor[rgb]{ 0,  0,  1}{7.077}  & 0.825  & \textcolor[rgb]{ 1,  0,  0}{1.829}  & 0.047  & 2.682  & \textcolor[rgb]{ 1,  0,  0}{0.549}  & 0.271  & \textcolor[rgb]{ 1,  0,  0}{9.726}  \\
    GANMcC & 6.967  & 0.760  & \textcolor[rgb]{ 0,  0,  1}{1.727}  & 0.047  & 2.694  & \textcolor[rgb]{ 0,  0,  1}{0.546}  & 0.271  & \textcolor[rgb]{ 1,  0,  0}{9.726}   \\
    UMF-CMGR & 6.641  & 0.744  & 1.641  & \textbf{0.033}  & 3.153 & 0.540  & \textcolor[rgb]{ 1,  0,  0}{0.401}  & 9.139  \\
    SwinFusion & 6.924  & \textbf{0.911}  & 1.675  & 0.059  & \textcolor[rgb]{ 1,  0,  0}{4.207}  & 0.501  & \textbf{0.493}  & 9.577  \\
    LRRNet & \textbf{7.127}  & \textcolor[rgb]{0,  0,  1}{0.835} & 1.604  & 0.051  & \textcolor[rgb]{ 0,  0,  1}{4.029}  & 0.476  & 0.347  & 9.457  \\
    FAFusion & 6.819  & 0.634 & 1.179  & 0.060  & \textbf{5.041}  & 0.364  & 0.346  & 9.362  \\
    Ours  & \textcolor[rgb]{ 1,  0,  0}{7.111}  & \textcolor[rgb]{ 1,  0,  0}{0.850}  & \textbf{1.850}  & \textcolor[rgb]{ 0,  0,  1}{0.044}  & 3.524  & \textbf{0.550}  & \textcolor[rgb]{ 0,  0,  1}{0.357}  & \textbf{9.735}  \\
    \midrule
    \multicolumn{1}{c}{\multirow{2}[4]{*}{Method}} & \multicolumn{8}{c}{$RoadScene$ Dataset} \\
\cmidrule{2-9}    \multicolumn{1}{c}{} & EN$\uparrow$  & VIF$\uparrow$ & SCD$\uparrow$ & MSE$\downarrow$ & AG$\uparrow$  & CC$\uparrow$  & $Q^{AB/F}$$\uparrow$& SD$\uparrow$ \\
    \midrule
    DenseFuse & 6.839  & 0.668  & 1.360  & \textbf{0.027}  & 3.742  & \textcolor[rgb]{ 1,  0,  0}{0.659}  & 0.363  & 9.573  \\
    RFN-Nest   & \textcolor[rgb]{1,  0,  0}{7.300}  & 0.660  & \textcolor[rgb]{ 1,  0,  0}{1.681}  & 0.043  & 3.228  & 0.645  & 0.281  & \textcolor[rgb]{ 0,  0,  1}{10.109}  \\
    GANMcC & 7.011  & 0.637  & 1.445  & 0.072  & 3.767  & \textcolor[rgb]{ 0,  0,  1}{0.646}  & 0.313  & 9.916  \\
    UMF-CMGR & 7.047  & \textcolor[rgb]{ 0,  0,  1}{0.733}  & 1.461  & \textcolor[rgb]{ 1,  0,  0}{0.029}  & 4.125  & 0.644  & \textbf{0.484}  & 9.821  \\
    SwinFusion & 6.922  & \textbf{0.780}  & \textcolor[rgb]{ 0,  0,  1}{1.591}  & 0.060  & \textcolor[rgb]{ 0,  0,  1}{4.491} & 0.624  & \textcolor[rgb]{ 1,  0,  0}{0.455}  & \textbf{10.474}  \\
    LRRNet & \textcolor[rgb]{ 0,  0,  1}{7.135}  & 0.701  & 1.569  & 0.068  & \textcolor[rgb]{ 1,  0,  0}{4.703} & 0.620  & 0.349  & 10.070  \\
    FAFusion & 6.991  & 0.559 & 0.798  & 0.061  & \textbf{5.694} & 0.464  & 0.371  & 10.088  \\
    Ours  & \textbf{7.370}  & \textcolor[rgb]{ 1,  0,  0}{0.757}  & \textbf{1.790}  & \textcolor[rgb]{ 0,  0,  1}{0.037}  & 4.472 & \textbf{0.662}  & \textcolor[rgb]{ 0,  0,  1}{0.401} & \textcolor[rgb]{ 1,  0,  0}{10.356}  \\
    \midrule
    \multicolumn{1}{c}{\multirow{2}[4]{*}{Method}} & \multicolumn{8}{c}{$M^3FD$ Dataset} \\
\cmidrule{2-9}    \multicolumn{1}{c}{} & EN$\uparrow$  & VIF$\uparrow$ & SCD$\uparrow$ & MSE$\downarrow$ & AG$\uparrow$  & CC$\uparrow$  & $Q^{AB/F}$$\uparrow$& SD$\uparrow$ \\
    \midrule
    DenseFuse & 6.698  & 0.762  & \textcolor[rgb]{ 0,  0,  1}{1.695}  & \textbf{0.032}  & 3.302  & \textbf{0.585}  & \textcolor[rgb]{ 1,  0,  0}{0.503}  & 9.021  \\
    RFN-Nest   & \textcolor[rgb]{ 0,  0,  1}{6.863}  & \textcolor[rgb]{ 0,  0,  1}{0.773}  & \textcolor[rgb]{ 1,  0,  0}{1.726}  & \textcolor[rgb]{ 0,  0,  1}{0.034}  & 2.855  & \textcolor[rgb]{ 1,  0,  0}{0.572}  & 0.405  & 9.232  \\
    GANMcC & 6.624  & 0.664  & 1.576  & 0.037  & 2.412  & \textcolor[rgb]{ 0,  0,  1}{0.571}  & 0.267  & 9.002  \\
    UMF-CMGR & 6.699  & 0.709  & 1.570  & \textbf{0.032}  & 2.929  & 0.546 & 0.397  & 9.144  \\
    SwinFusion & 6.790  & \textbf{0.877}  & 1.562  & 0.060  & \textcolor[rgb]{ 1,  0,  0}{4.615}  & 0.521  & \textbf{0.597}  & \textbf{9.859}  \\
    LRRNet & 6.437  & 0.719  & 1.463  & 0.039  & 3.600 & 0.542  & \textcolor[rgb]{ 0,  0,  1}{0.497}  & 9.307  \\
    FAFusion & \textcolor[rgb]{ 1,  0,  0}{6.895} & 0.689 & 1.254  & 0.051  & \textbf{5.799} & 0.439  & 0.422  & \textcolor[rgb]{ 0,  0,  1}{9.692}  \\
    Ours  & \textbf{6.909}  & \textcolor[rgb]{ 1,  0,  0}{0.816}  & \textbf{1.778}  & \textcolor[rgb]{ 1,  0,  0}{0.033}  & \textcolor[rgb]{ 0,  0,  1}{3.602}  & \textbf{0.585}  & 0.487  & \textcolor[rgb]{ 1,  0,  0}{9.737}  \\
    \bottomrule
    \end{tabular}%
    }
    }
  \label{tab:addlabel}%
\end{table}%

%------------------------------------------------------------------------- 
\subsubsection{Quantitative Evaluation}

Quantitative comparison results listed in Table \ref{tab:addlabel} show that our method achieves comparable results on the \emph{TNO} dataset. From Table \ref{tab:addlabel} we can see that our method MDA achieves the largest average value in SCD, CC and SD, and obtain comparable statistics values on EN, VIF, MSE, AG and $ Q^{AB/F} $. For EN and SD, images with higher contrast usually achieve better performance, such as RFN-Nest and GANMcC hihglight the human in the fourth and fifth columns. However, our MDA expects to preserve pixel intensity and texture detail simultaneously, and its results retain comparable contrast while achieving more realistic edges, which satisfy human visual perception better. SCD, CC and AG show that our results maintain rich edge information transferred from the input image pair. These eight statistics values demonstrate that our results preserve the largest amount of complementary information transferred from the input image pairs with fewer artifacts, which satisfies the human visual system.

%------------------------------------------------------------------------- 
\begin{figure*}[!t]
\centering
\includegraphics[width=5.5in]{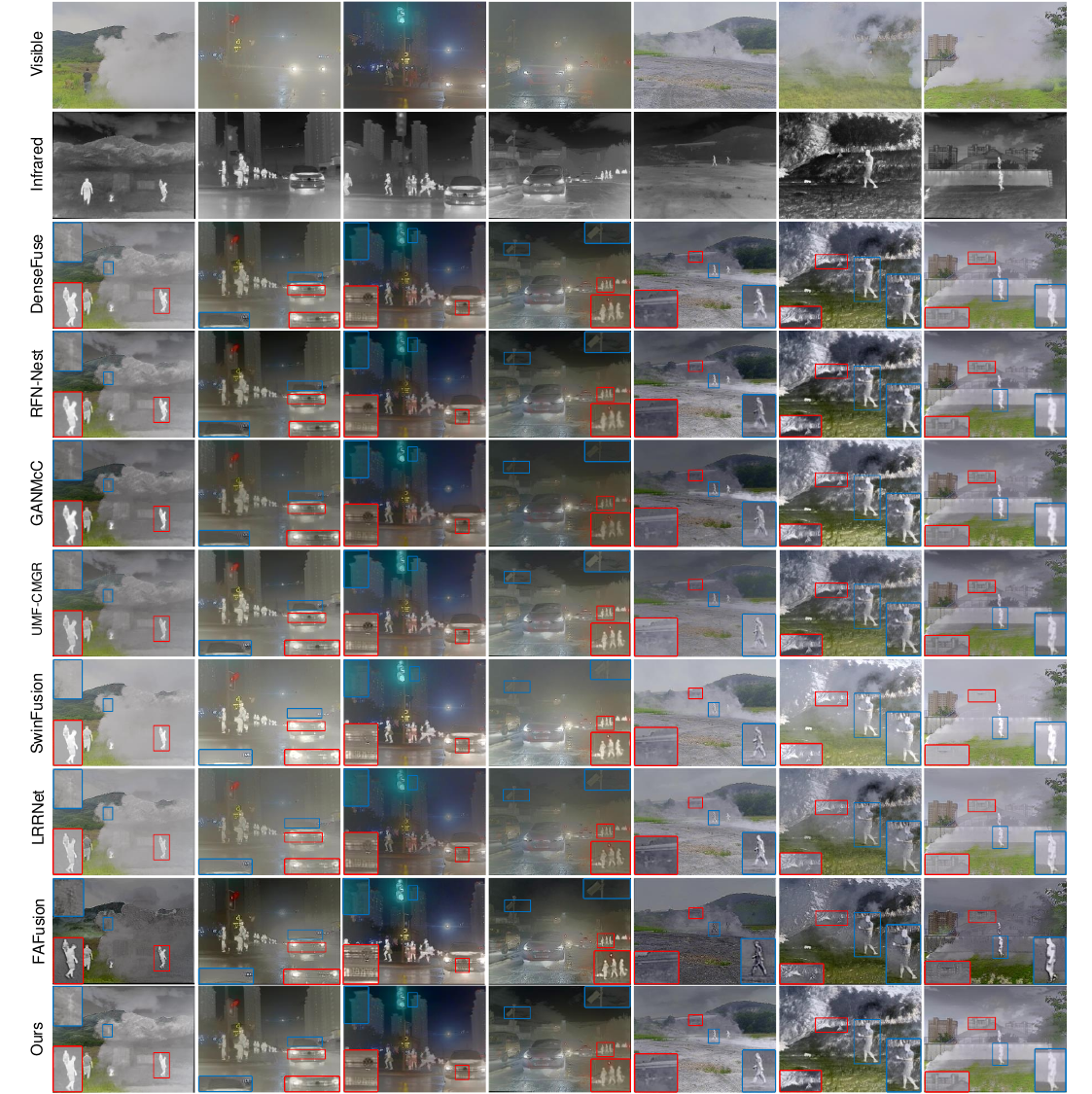}
\caption{
Qualitative comparison of our method with 7 state-of-the-art models on seven infrared and visible image pairs in the $M^3FD$ dataset. The first and second rows are visible and infrared images, respectively. From the third to the last rows are fusion results of DenseFuse\cite{li2018densefuse}, RFN-Nest\cite{li2021rfn}, GANMcC\cite{ma2020ganmcc}, UMF-CMGR\cite{diunsupervised}, SwinFusion\cite{ma2022swinfusion}, LRRNet\cite{li2023lrrnet}, FAFusion\cite{xiao2024fafusion} and our MDA.
}
\label{fig:model_comparison_m3fd}
\end{figure*}

\subsection{Results on RoadScene Dataset}
We further validate the performance of our method on the \emph{RoadScene} dataset, which consists of color visible and grayscale infrared image pairs with better imaging quality.

\subsubsection{Qualitative Evaluation}
Fig. \ref{fig:model_comparison_roadscene} exhibits qualitative comparison results of different models on the \emph{RoadScene} dataset. We select 7 image pairs to illustrate fusion performance for the self-driving scene, including day and night. Our fusion results exhibit minimal artifacts compared with the other seven state-of-the-art methods, such as the car with a clear structure in the red box of the first and third columns. None of the pedestrian targets are salient enough in visible images of the last four columns, and almost all the fusion results succeed in focusing on the pedestrian targets. Among them, the fusion performance of UMF-CMGR and our MDA keep the most similar style to the infrared image. However, as with results presented in the \emph{TNO} dataset, our results retain both thermal and detail information in the red box region, which indicates that our method is capable of adaptively measuring and integrating complementary information for global and local areas. Additionally, the texture of the sky and branches in the third and fourth columns are preserved, meanwhile the thermal target maintains a high contrast with the surrounding environment in our MDA method, suggesting that the intensity and detail information from the infrared images are adaptively measured.

\subsubsection{Quantitative Evaluation}
Quantitative comparison results are listed in Table \ref{tab:addlabel}, and our results are the best for EN, SCD and CC, as the quantitative results in the \emph{TNO} dataset, which indicates that our fusion results adaptively preserve the edge and detail information of input image pairs. Specifically, the mean value of SCD and CC in \emph{TNO} and \emph{RoadScene} datasets is higher than other models, corresponding to the richest transferred information from the input image pairs. For VIF and SD, our results are not optimal but not too far from the best results, which can be considered as a trade-off of adding complementary information from the visible image to the higher contrast infrared image. In summary, our fusion results generate rich texture detail information meanwhile the preservation of saliency thermal targets is comparable.

\begin{figure}
\centering
\includegraphics[width=5.5in]{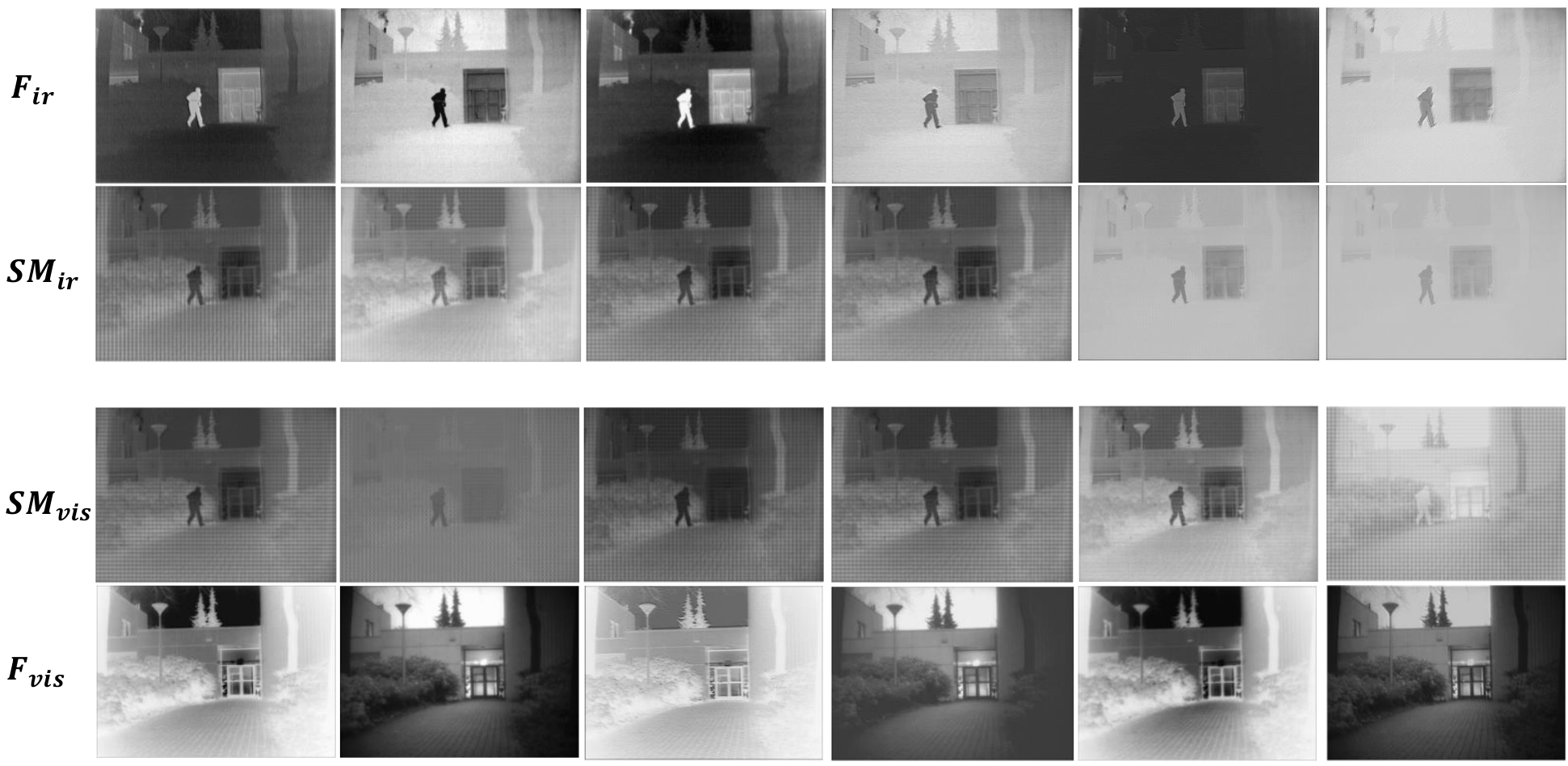}
\caption{
Visualization of spatial attention map. From left to right are six channels selected from the input feature maps and spatial attention maps of the dual attention fusion block.
}
\label{fig:attention}
\end{figure}

\subsection{Results on $M^3FD$ Dataset}
We have also tested our method on the $M^3FD$ dataset, which consists of high-resolution color visible and grayscale infrared image pairs, to explore the fusion performance in night and bad weather scenarios.

\subsubsection{Qualitative Evaluation}
Fig. \ref{fig:model_comparison_m3fd} exhibits qualitative comparison results of different models on the $M^3FD$ dataset. We select 7 image pairs to exhibit fusion performance comparison for poor visual conditions, like rainy nights and smog. Due to bad weather conditions and smog, the thermal targets and texture information of visible images are obscured by the smog. Almost all the methods are able to highlight the thermal target in the infrared images, but some of them fail to preserve detail information behind the smog, such as the SwinFusion. Our method preserves the high contrast between the target and background of the infrared image, while the complex texture only can be seen in the infrared image is also preserved, such as the mountain and buildings behind the smog in the first, fifth and last columns.

The fusion results of our MDA keep a similar intensity style with infrared images, such as the fused images from the second to fourth columns. Visible images are full of fog, which causes halo effect around the light and low contrast. Our fusion results preserve high contrast from the infrared images and halo effect is reduced, like the sharp traffic light in the third column.

\subsubsection{Quantitative Evaluation}
Quantitative comparison results are listed in Table \ref{tab:addlabel}, and our results are the best for EN, SCD and CC, meanwhile the VIF, MSE, AG and SD also achieve comparative results. The largest value of CC and SCD indicate that our fusion results are highly correlated with the original image pairs, and the second largest SD and AG values of our method indicate that our method keeps comparative intensity contrast and texture detail information simultaneously. 

\begin{table*}[htbp]
  \centering
  \caption{Quantitative results of ablation study on the \emph{TNO} dataset. The best performance are shown in \textbf{bold}, and the second and third best performance are shown in \textcolor[rgb]{ 1,  0,  0}{red} and \textcolor[rgb]{ 0,  0,  1}{blue}, respectively.}
  \resizebox{\textwidth}{!}{
   \setlength{\tabcolsep}{3mm}{
    \begin{tabular}{l|cccccccc}
     \toprule
\cmidrule{1-9}    \multicolumn{1}{c}{Method} & EN $\uparrow$ & VIF $\uparrow$ & SCD $\uparrow$ & MSE $\downarrow$ & AG $\uparrow$ & CC $\uparrow$  & $Q^{AB/F}$ $\uparrow$ & SD $\uparrow$ \\
    \midrule
    summation   & \textcolor[rgb]{ 1,  0,  0}{7.125} & \textcolor[rgb]{ 1,  0,  0}{0.845} & 1.617 & 0.060 & 2.728 & 0.467 & \textcolor[rgb]{ 0,  0,  1}{0.242} & 9.271 \\
    concatenation & 7.082 & \textcolor[rgb]{ 0,  0,  1}{0.820} & \textcolor[rgb]{ 1,  0,  0}{1.816} & \textcolor[rgb]{ 0,  0,  1}{0.047} & \textcolor[rgb]{ 0,  0,  1}{3.159} & \textcolor[rgb]{ 1,  0,  0}{0.543} & \textcolor[rgb]{ 1,  0,  0}{0.316} & \textbf{10.072} \\
    $w/o$ adaptive $L_{image}$ & 7.080 & 0.763 & \textcolor[rgb]{ 0,  0,  1}{1.774} & \textbf{0.043} & 2.231 & \textcolor[rgb]{ 0,  0,  1}{0.529} & 0.167 & 9.676 \\
    $w/o$ adaptive $L_{patch}$ & \textbf{7.149} & 0.712 & 1.760 & \textcolor[rgb]{ 1,  0,  0}{0.044} & 1.962 & 0.521 & 0.132 & \textcolor[rgb]{ 1,  0,  0}{9.737} \\
    $L_{pixel}$: pooling 1-3 & 6.940 & 0.783 & 1.559 & 0.054 & \textcolor[rgb]{ 1,  0,  0}{3.476} & 0.467 & 0.176 & 9.194 \\
    $L_{pixel}$: pooling 1-4 &  \textcolor[rgb]{ 0,  0,  1}{7.120} & 0.746 & 1.760 & \textbf{0.043} & 2.105 & 0.522 & 0.146 & 9.691 \\
    Ours  &  7.111  & \textbf{0.850}  & \textbf{1.850}  & \textcolor[rgb]{ 1,  0,  0}{0.044}  & \textbf{3.524}  & \textbf{0.550}  & \textbf{0.357}  &  \textcolor[rgb]{ 0,  0,  1}{9.735} \\
    \bottomrule
    \end{tabular}%
    }}
  \label{tab:headings}%
\end{table*}%

%------------------------------------------------------------------------- 
\subsection{Ablation Study}

In our study, we measure the complementary information via the statistical methods and features extracted by pre-trained VGG-16. In this section, we perform the ablation study to validate the effectiveness of the complementary information measurement, information aggregation structure and hyperparameters in the network.

\subsubsection{Evaluation of the fusion block}
As we discussed in the previous section, dual attention fusion block combines spatial and channel attention modules to integrate complementary information and suppress redundant information. Specifically, the spatial attention module exploits the inter-spatial relationship between the infrared and visible image. Based on the above consideration, we visualize the spatial attention map $ {SM}_{ir} $, $ {SM}_{vis} $ before multiplying with $ F_{ir} $, $ F_{vis} $ to validate the spatial attention map is capable of focusing on regions contains vital information. In Fig. \ref{fig:attention}, we can see that spatial attention maps $ {SM}_{ir} $ and its corresponding inputs $ F_{ir} $ are highly correlated. In this example image, the vital information for infrared feature maps $ F_{ir} $ are regions including person and door, which are not significant in the visible feature maps $ F_{vis} $. However, these regions in visible attention maps are highlighted, which means the information between two modalities has been effectively interacted with and integrated via the dual attention fusion block. Likewise, the infrared attention maps contain rich detail texture information in the region of plants, where the input infrared image has poor texture in the same region.

\begin{figure}
\centering
\includegraphics[width=4.5in]{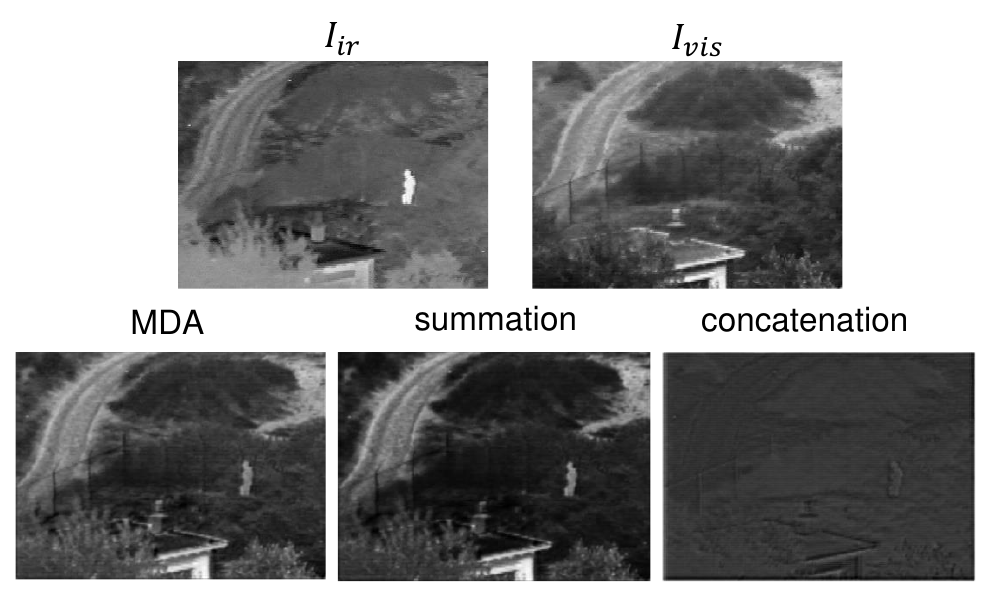}
\caption{
Illustration of fusion strategy results. The first row indicates the input infrared and visible image pair, and three fusion results in the second row are our MDA,  replacing fusion block with the summation and concatenation, respectively.
}
\label{fig:ablation_fusion}
\end{figure}

\begin{figure}
\centering
\includegraphics[width=4.5in]{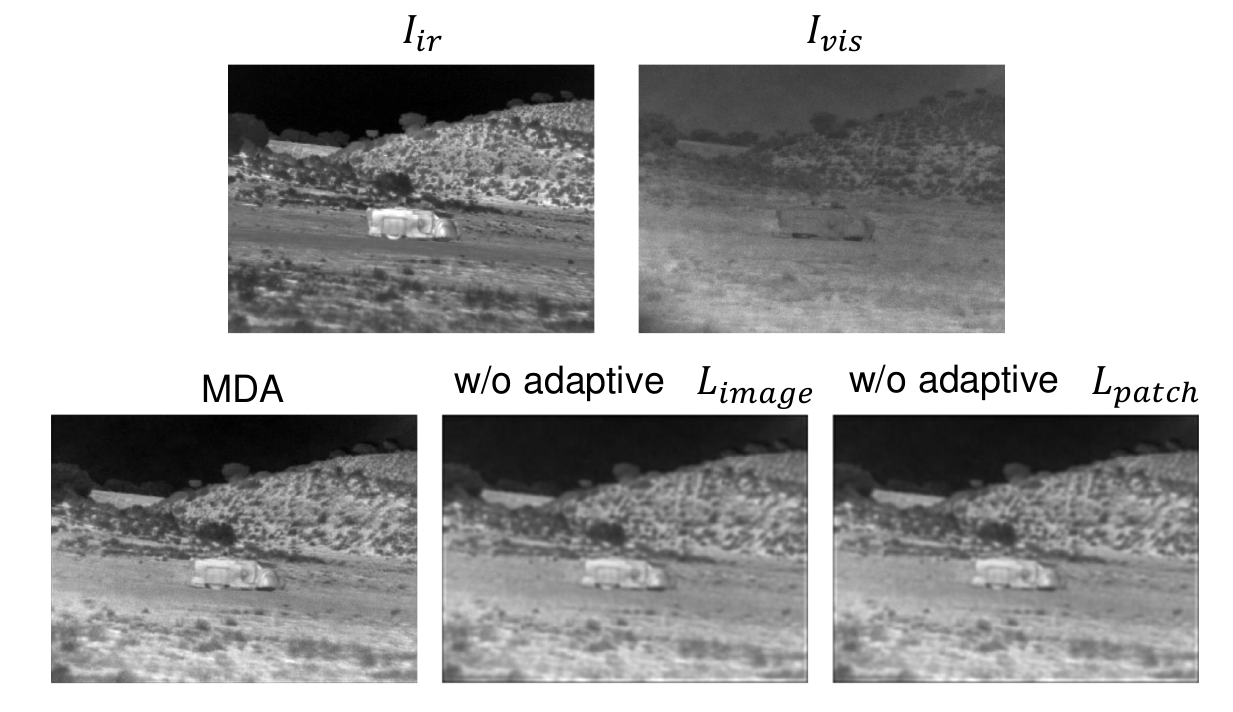}
\caption{
Illustration of fusion results of fixed weight in $ L_{image} $ and $ L_{patch} $. The first row indicates the input infrared and visible image pair, and the second row is our MDA model, fixing weight coefficients to 0.5 of $L_{image}$ and fixing weight coefficients to 0.5 of $L_{patch}$, respectively.
}
\label{fig:ablation_loss}
\end{figure}

To exploit the effectiveness of the dual attention block, we compare our fusion method with two other fusion ways, which are fusion via summation and concatenation. The qualitative results on the \emph{TNO} dataset are shown in Fig. \ref{fig:ablation_fusion}. As we can see, our MDA provides the best balance of contrast and detail. The summation fusion results achieve the highest contrast, but a lot of the detailed information in the regions with low pixel intensity is lost. 

\begin{figure}
\centering
\includegraphics[width=4.5in]{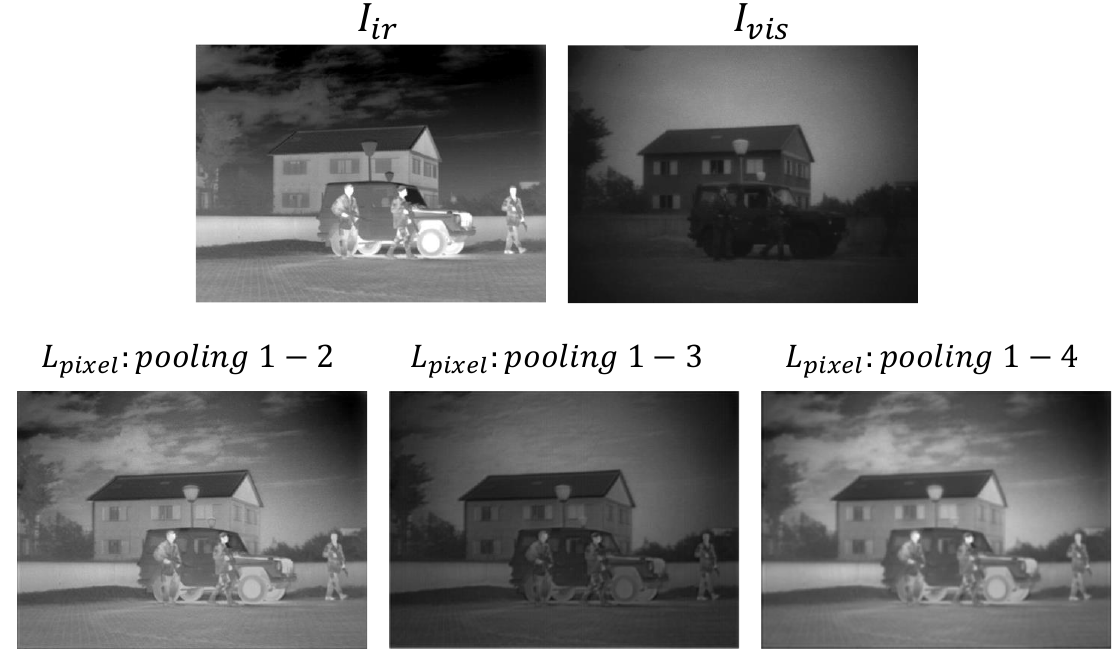}
\caption{
Results of different layers of VGG-16 feature maps used to estimate adaptive weight coefficients. The first row indicates the input infrared and visible image pair, and the second row is our MDA model, using feature maps before the 1,2,3 pooling layers of VGG-16 and feature maps before the 1,2,3,4 pooling layers of VGG-16, respectively.
}
\label{fig:ablation_vgg}
\end{figure}

\subsubsection{Evaluation of the adaptive weight}
To validate the effectiveness of our adaptive weight in measuring the complementary information, we perform two ablation experiments for fixed weight and different VGG-16 feature depths in weight generation. We perform the first experiment where $ \omega^{ir} $ and $ \omega^{vis} $ of $L_{image}$ or $ L_{patch} $ are set to 0.5, respectively. The qualitative comparison results on the \emph{TNO} dataset are shown in Fig. \ref{fig:ablation_loss}. The first fusion result is generated by our MDA method, which adopts adaptive weight coefficient in both $L_{image}$ and $L_{patch}$, and the last two fusion results are only fixing the weight of $L_{image}$ and $ L_{patch} $, respectively. We can see that without the adaptive weight generated by the method proposed in this paper, fusion results of both fixed weights lack accurate detailed information.

Additionally, as we discussed in Section III, feature maps in the deep layers of VGG-16 contain poor information in terms of our data, thus we illustrate the fusion results using weights generated from VGG-16 feature maps of different depths, as shown in Fig. \ref{fig:ablation_vgg}. The qualitative results validate the analysis in Section III, in which feature maps in the deep layers of VGG-16 barely convey intensity information, consistent with the thermal target information lost in the fusion results of $L_{pixel}$: pooling 1-3 and $L_{pixle}$: pooling 1-4.

\subsubsection{Quantitative ablation experiments}
Table \ref{tab:headings} reports the quantitative ablation experiments. The average values of ablation experiment results are consistent with the comparison experiment with state-of-the-art models, i.e. the average value of VIF, SCD, CC, AG and $Q^{AB/F}$ of our MDA are highest compared with others, which indicates that our MDA preserves the best detail and edge information from the input image pairs. Without using the adaptive weight in the $ L_{image}$ and $L_{patch}$, the value of AG has dropped, which indicates that the global contrast information obtains insufficient preservation, and so does the fusion via the summation and concatenation methods.

\subsubsection{Evaluation of the $\alpha$, $\beta$, $\gamma$ and size of the $W$}
We analyze the influence of hyperparameters  $\alpha$, $\beta$, $\gamma$ and size of the sliding window $W$ in the loss function and the quantitative result on the \emph{TNO} dataset is shown in Table \ref{tab:hyperpara}. The $\alpha$, $\beta$ and $\gamma$ are the weight controlling $L_{feature}$, $L_{style}$ and $L_{patch}$ in the loss function separately. $W$ controlls the size of the sliding window of the $L_{patch}$. We choose fourteen sets of hyperparameter values to test the influence of the different hyperparameters in the loss function, which are  $\alpha=1e-5$, $\alpha=1e-6$, $\alpha=1e-10$, $\alpha=1e-11$, $\beta=1e3$, $\beta=1e4$, $\beta=1e6$, $\beta=1e7$, $\gamma=5$, $\gamma=3$, $\gamma=1.5$, $\gamma=1$,  $W=15$,  $W=23$. The comparison result demonstrates that AG and $Q^{AB/F}$ exhibit relatively high sensitivity to hyperparameters, which correspond to the retention of edge information. The variation of metrics changes relatively small when the $\alpha$ and $\beta$ change, which means the fusion result is less influenced by the $L_{feature}$ and $L_{style}$ compared with $L_{image}$. On the contrary, the variation of metrics changes more significantly when the $\gamma$ changes, especially the value of the VIF,  AG and $Q^{AB/F}$, which indicates the balance between the global and local complementary information measurement would change the preservation of the intensity and detail information.

\begin{table*}[htbp]
  \centering
  \caption{Quantitative results of the $\alpha$, $\beta$, $\gamma$ and size of the $W$ on the \emph{TNO} dataset. The best performance are shown in \textbf{bold}, and the second and third best performance are shown in \textcolor[rgb]{ 1,  0,  0}{red} and \textcolor[rgb]{ 0,  0,  1}{blue}, respectively.}
  \renewcommand{\arraystretch}{0.7}
  \resizebox{\textwidth}{!}{
   \setlength{\tabcolsep}{3mm}{
    \begin{tabular}{l|cccccccc}
     \toprule
\cmidrule{1-9}    \multicolumn{1}{c}{Method} & EN $\uparrow$ & VIF $\uparrow$ & SCD $\uparrow$ & MSE $\downarrow$ & AG $\uparrow$ & CC $\uparrow$  & $Q^{AB/F}$ $\uparrow$ & SD $\uparrow$ \\
    \midrule
    $\alpha=1e-5$ & 6.847 & 0.776 & 1.652 & \textcolor[rgb]{ 1,  0,  0}{0.045} & 2.950 & 0.510 & 0.253 & 9.285 \\
    $\alpha=1e-6$ & 7.170 & 0.645 & 1.670 & 0.048 & 2.074 & 0.527 & 0.229 & 9.755 \\
    $\alpha=1e-10$  & \textbf{7.302} & \textcolor[rgb]{ 0,  0,  1}{0.833} & \textcolor[rgb]{ 1,  0,  0}{1.825} & 0.049 & \textcolor[rgb]{ 0,  0,  1}{2.974} & \textcolor[rgb]{ 1,  0,  0}{0.539} & \textcolor[rgb]{ 0,  0,  1}{0.287} & \textbf{10.040} \\
    $\alpha=1e-11$ & 7.208 & 0.832 & \textcolor[rgb]{ 0,  0,  1}{1.818} & \textcolor[rgb]{ 0,  0,  1}{0.047} & 2.662 & \textcolor[rgb]{ 0,  0,  1}{0.534} & 0.264 & 9.806 \\
     \midrule
    $\beta=1e7$ & 7.245 & 0.759 & 1.766 & 0.049 & 2.349 & 0.521 & 0.259 & \textcolor[rgb]{ 1,  0,  0}{9.935} \\
    $\beta=1e6$ & 7.277 & 0.785 & 1.768 & 0.048 & 2.378 & 0.515 & 0.235 & \textcolor[rgb]{ 0,  0,  1}{9.855} \\
    $\beta=1e4$ & 7.282 & 0.789 & 1.765 & 0.048 & 2.379 & 0.513 & 0.234 & 9.825 \\
    $\beta=1e3$ & 7.201 & \textbf{0.878} & 1.797 & \textcolor[rgb]{ 0,  0,  1}{0.047} & 2.931 & 0.522 & 0.240 & 9.774 \\
     \midrule
     $\gamma=5$ & 7.289 & 0.802 & 1.770 & 0.049 & 2.210 & 0.515 & 0.203 & 9.823 \\
    $\gamma=2.5$ & 7.281 & 0.796 & 1.769 & 0.049 & \textcolor[rgb]{ 1,  0,  0}{3.102} & 0.514 & \textcolor[rgb]{ 1,  0,  0}{0.307} & 9.833 \\
    $\gamma=1.5$ & 7.281 & 0.785 & 1.763 & 0.048 & 2.965 & 0.512 & 0.279 & 9.826 \\
    $\gamma=1$ & 7.289 & 0.777 & 1.751 & 0.048 & 2.235 & 0.509 & 0.188 & 9.803 \\
     \midrule
     $W=23$ & \textcolor[rgb]{ 1,  0,  0}{7.291} & 0.787 & 1.760 & 0.048 & 2.164 & 0.512 & 0.197 & 9.819 \\
    $W=15$ & \textcolor[rgb]{ 0,  0,  1}{7.290} & 0.786 & 1.760 & 0.048 & 2.165 & 0.511 & 0.223 & 9.820 \\
    Ours  &  7.111  & \textcolor[rgb]{ 1,  0,  0}{0.850}  & \textbf{1.850}  & \textbf{0.044}  & \textbf{3.524}  & \textbf{0.550}  & \textbf{0.357}  & 9.735 \\
    \bottomrule
    \end{tabular}%
    }}
  \label{tab:hyperpara}%
\end{table*}%

%===========================================================
\section{Conclusion}
In this paper, we propose a multi-scale framework for infrared and visible image fusion. We construct the residual downsample block and residual reconstruction block for multi-scale feature extraction and image reconstruction. Dual attention fusion block is applied to aggregate the complementary information from infrared and visible branches based on the spatial and channel attention module. Lastly, the information retention degree is obtained by the statistic methods to control weights of the intensity and gradient items in the loss function, which are used to constrain the information transferred from the inputs. We validate the effectiveness of our method on the \emph{TNO}, \emph{RoadScene} and $M^3FD$ datasets. Qualitative and quantitative results show that our method achieves comparable results compared with the state-of-the-art methods and best satisfies human visual perception.

\section{Acknowledgements}
This work is supported by the National Natural Science Foundation of China under Grants U21A20514.

\section{Declaration of interests}
The authors declare that they have no known competing financial interests or personal relationships that could have appeared to influence the work reported in this paper.

%% Loading bibliography style file
%\bibliographystyle{model1-num-names}
%\bibliographystyle{elsarticle-num-names}
\bibliographystyle{elsarticle-num}

% Loading bibliography database
\bibliography{cas-refs}

\end{document}